\documentclass[a4paper,fleqn]{cas-dc}

\usepackage[numbers]{natbib}
\usepackage{color}

\usepackage{booktabs}

\def\tsc#1{\csdef{#1}{\textsc{\lowercase{#1}}\xspace}}
\tsc{WGM}
\tsc{QE}
\tsc{EP}
\tsc{PMS}
\tsc{BEC}
\tsc{DE}
\begin{document}
\let\WriteBookmarks\relax
\def\floatpagepagefraction{1}
\def\textpagefraction{.001}
\shorttitle{Discovery and inversion of the viscoelastic wave equation in inhomogeneous media}
\shortauthors{Su Chen et~al.}

\title [mode = title]{Discovery and inversion of the viscoelastic wave equation in inhomogeneous media}

\author[1]{Su Chen}
\credit{Conceptualization, Investigation, Methodology, Writing-Review\& Editing}

\author[1]{Yi Ding}[orcid=0000-0003-3443-9313]
\ead{dingyi18@mails.ucas.edu.cn}
\cormark[1]
\credit{Methodology, Software, Visualization, Writing-Original Draft, Writing-Review\& Editing}

\author[3]{Hiroe Miyake}
\credit{Formal analysis, Supervision, Validation, Writing-Review\& Editing}

\author[1,2]{Xiaojun Li}
\credit{Resources, Supervision, Writing-Review\& Editing}

\affiliation[1]{organization={Key Laboratory of Urban Security and Disaster Engineering of the Ministry of Education, Beijing University of Technology},
                city={Beijing},
%               citysep={}, % Uncomment if no comma needed between city and postcode
                postcode={100124},
                country={China}}

\affiliation[2]{organization={Institute of Geophysics, China Earthquake Administration},
                city={Beijing},
                postcode={100081},
                country={China}}

\affiliation[3]{organization={Earthquake Research Institute, University of Tokyo},
                city={Tokyo},
                postcode={},
                postcodesep={},
                country={Japan}}

\cortext[cor1]{Corresponding author}

\begin{abstract}
In scientific machine learning, the task of identifying partial differential equations accurately from sparse and noisy data poses a significant challenge. Current sparse regression methods may identify inaccurate equations on sparse and noisy datasets and are not suitable for varying coefficients. To address this issue, we propose a hybrid framework that combines two alternating direction optimization phases: discovery and embedding. The discovery phase employs current well-developed sparse regression techniques to preliminarily identify governing equations from observations. The embedding phase implements a recurrent convolutional neural network (RCNN), enabling efficient processes for time-space iterations involved in discretized forms of wave equation. The RCNN model further optimizes the imperfect sparse regression results to obtain more accurate functional terms and coefficients. Through alternating update of discovery-embedding phases, essential physical equations can be robustly identified from noisy and low-resolution measurements. To assess the performance of proposed framework, numerical experiments are conducted on various scenarios involving wave equation in elastic/viscoelastic and homogeneous/inhomogeneous media. The results demonstrate that the proposed method exhibits excellent robustness and accuracy, even when faced with high levels of noise and limited data availability in both spatial and temporal domains.
\end{abstract}

\begin{highlights}
\item Propose an alternating optimization discovery-embedding framework for discovery and inversion of the viscoelastic wave equation in inhomogeneous media.
\item Establish a conceptual comparison between viscoelastic wave physics system and standard recurrent neural network.
\item Embedding homogeneous Dirichlet, Neumann, and absorbing boundary conditions into the optimized recurrent convolutional neural network.
\end{highlights}


\maketitle

\section{Introduction}
Extracting partial differential equations (PDEs) from measurement data remains critical for modeling, simulation, and understanding the dynamical patterns and complex spatiotemporal behaviors in nature \cite{Xu_2023_DiscoveryPartialDifferential, Yu_2024_LearningDynamicalSystems}. While observations can often be described by derived physical laws (e.g., the wave equations), there are cases where observations do not conform to these laws or have not yet been physically described. And as the complexity and nonlinearity of the system increase, deriving the governing equations becomes increasingly challenging. In recent decades, discovering causal relations and underlying governing laws from data have emerged as exciting fields of research that promise advancing science \cite{Camps-Valls_2023_DiscoveringCausalRelations}.

Recently, thanks to the advances in observational datasets and the thriving machine-learning community, there has been significant progress in distilling physical laws and governing equations from data \cite{Bongard_2007_AutomatedReverseEngineering, Schmidt_2009_DistillingFreeFormNatural}. Current methods are mainly divided into closed and expandable library methods based on the library construction method. Expandable library methods do not need to predetermine an overcomplete library, they only need a randomly generated incomplete initial library to generate various unknown combinations by introducing methods such as genetic algorithms \cite{Xu_2020_DLGAPDEDiscoveryPDEs, Chen_2022_SymbolicGeneticAlgorithm, Makke_2024_InterpretableScientificDiscovery}. Cheng and Alkhalifah \cite{Cheng_2023_RobustDataDriven, Cheng_2024_DiscoveryPhysicallyInterpretable} combine neural networks and genetic algorithms for discovering the acoustic wave equation. One of the most popular closed library approaches for identifying physical laws is sparse identification of nonlinear Dynamics (SINDy) \cite{Brunton_2016_DiscoveringGoverningEquations}, which introduces sparse linear regression into the discovery of parsimonious governing equations from a library of predefined candidate functions. Since then, the technique of sparse linear regression has drawn tremendous attention in the past few years, leading to variant applications across a range of scientific disciplines \cite{Loiseau_2018_SparseReducedorderModelling, Loiseau_2018_ConstrainedSparseGalerkin, Stender_2019_RecoveryDifferentialEquations}. Moreover, the sparsity-promoting paradigm, such as Sequential Threshold Ridge regression (STRidge), has been expanded to encompass more general PDE discovery. This involves augmenting the library of candidate functions by incorporating spatial partial derivative terms \cite{Rudy_2017_DatadrivenDiscoveryPartial}.

While previous research endeavors have achieved considerable success in data-driven PDE discovery, the standard sparse representation-based methods may yield inaccurate results as sparsity and noise levels escalate, posing a significant challenge in numerous real-world applications. Recently, some studies have aimed to enhance the ability of sparse regression to identify PDEs in noisy and sparse observation scenarios \cite{Messenger_2021_WeakSINDyPartial, Sashidhar_2022_BaggingOptimizedDynamic, Fasel_2022_EnsembleSINDyRobustSparse}. On the other hand, efforts have focused on encoding prior physical knowledge into network architectures to facilitate the learning of spatiotemporal dynamics in scenarios with sparse data \cite{Chen_2021_PhysicsinformedLearningGoverning, Rao_2023_EncodingPhysicsLearn}. Recent research on modeling physical systems using neural networks can be divided into two categories: continuous and discrete approaches. An typical work in continuous learning is the physics-informed neural networks based on fully connected neural networks \cite{Raissi_2019_PhysicsinformedNeuralNetworks}, which have been used for forward \cite{Song_2021_SolvingFrequencydomainAcoustic, Ding_2023_SelfadaptivePhysicsdrivenDeep, Ding_2023_PhysicsconstrainedNeuralNetworks, Ren_2024_SeismicNetPhysicsinformedNeural} and inverse \cite{Rasht-Behesht_2022_PhysicsInformedNeuralNetworks, Zhang_2023_SeismicInversionBased} analysis of wave equations. On the other hand, due to the lightweight architecture and hard-embedded characteristics of boundary conditions, physics-informed approaches based on convolutional neural network (CNN) have been used for PDE system modeling in discrete learning \cite{Gao_2021_PhyGeoNetPhysicsinformedGeometryadaptive, Qu_2022_LearningTimedependentPDEs, Ren_2022_PhyCRNetPhysicsinformedConvolutionalrecurrent}. However, due to limitations in computational efficiency \cite{McGreivy_2024_WeakBaselinesReporting} and spectral bias \cite{Tancik_2020_FourierFeaturesLet, Wang_2021_EigenvectorBiasFourier}, learning the complex dynamics of wave propagation with neural networks remains challenging. Recently, by leveraging the inherent connection between temporal-spatial stepping processes and recurrent neural network (RNN) as well as convolutional layers (CLs), a recurrent convolutional neural network (RCNN) has been proposed for efficiently full-wave electromagnetic \cite{Guo_2023_ElectromagneticModelingUsing} modeling and mechanical wave modeling \cite{Ji_2024_EfficientPlatformNumerical} on GPU platforms. The weight parameters in RCNN are directly derived from finite difference time domain (FDTD) formulas, enabling efficient processes for time-space iterations involved in discretized forms of PDEs.

Sparse regression is inherently constrained to providing a scalar identification of coefficients, which limits its applicability in inverse problems involving spatially varying parameters. Cheng and Alkhalifah (2024) assume that velocity is measurable and develop a framework that simultaneously encodes velocity information alongside candidate function terms to derive the acoustic wave equations for both homogeneous and heterogeneous media \cite{Cheng_2024_DiscoveryPhysicallyInterpretable}. However, in geophysical exploration, obtaining complete velocity information poses significant challenges. Consequently, our objective is to tackle the identification and inverse problems associated with PDEs characterized by spatially varying parameters. To this end, we propose an innovative computational framework aimed at uncovering the viscoelastic wave equations of heterogeneous media from sparse and noisy measurement data. Our approach integrates sparse regression with an optimized RCNN model. The primary contributions of this paper are threefold:
\begin{enumerate}
\itemsep=0pt
\item We propose an alternating optimization discovery-embedding framework. Through sparse regression in the discovery phase, we identify essential equation terms and establish reasonable initial coefficients. Subsequently, the RCNN model in the embedding phase can filter equation terms and optimize coefficients, refining the discovered system further.
\item By utilizing the inherent connection between the time-marching process and RNN, we established a conceptual comparison between viscoelastic wave physics system and standard RNN.
\item We use a series of convolution kernels (filters) in CLs to approximate the discretization scheme of finite difference (FD) operators. Different boundary conditions are embedded into the FD-based filters, including homogeneous Dirichlet, Neumann, and absorbing boundary conditions.
\end{enumerate}

To convey this idea, we structure the rest of this paper as follows. The problem statement is defined in Section \ref{sec:problem-statement}. The concept of sparse regression and the limitations of its application are discussed in Section \ref{sec:sparse-discovery}. In Section \ref{sec:recurr-conv-neur}, we introduced the theory and optimization details of the RCNN model. Section \ref{sec:bound-cond-sett} describes different boundary conditions embedded in the RCNN model. In Section \ref{sec:numer-exper}, we implement a number of numerical experiments and show the results to evaluate the performance of our proposed approach. Sections \ref{sec:discussion} and \ref{sec:conclusion} finally provides a discussion and concluding remarks.

\section{Methodology}
\subsection{Problem statement}\label{sec:problem-statement}
We aim to devise a learning paradigm to solve the inverse PDEs identification problem. In this work, we consider the general form of a wave equation consisting of:
\begin{equation}
\label{eq:1}
u_{t t}=\mathcal{F}\left[u, u^{2}, \cdots, \nabla_{x} u, \nabla_{x}^{2} u, \nabla_{x} u \cdot u, \cdots, u_{t}, \nabla_{x} u \cdot u_{t} \cdots\right],
\end{equation}
where $u_{t}, u_{tt}$ are the first and second-order temporal derivatives of the displacement $u(x, t)$. $\mathcal{F}\left[\cdot\right]$ is a composite function describing the right-hand side (RHS) of PDEs, which involving combinations of $u$ and its spatial or temporal derivatives.

Our objective is to identify the closed form of $\mathcal{F}\left[\cdot\right]$ from available spatiotemporal measurements which are assumed to be incomplete, scarce, and noise-laden observations common in the real world. The task of identifying function $\mathcal{F}\left[\cdot\right]$ comprises two components. It involves the identification of suitable functional terms from an overcomplete candidate library. Secondly, it entails the determination of the corresponding nonzero coefficient vector, which may exhibit spatial variations, owing to the heterogeneity of the medium properties. Fortunately, there exists a parsimonious form for canonical PDEs (including the wave equation), wherein the RHS has only a limited number of terms in an active state.

We consider a one-dimensional (1D) wave equation containing a viscous term as shown in the following equation \cite{Idriss_1968_SeismicResponseHorizontal}:
\begin{equation}
\label{eq:2}
\frac{\partial^{2} u}{\partial t^{2}}=c^{2} \frac{\partial^{2} u}{\partial x^{2}}+\eta \frac{\partial u}{\partial t},
\end{equation}
where we assume that the body force is absent, $c$ is the spatially varying wave velocity, $\eta$ is defined as the viscous factor. The wavefield is excited by Ricker wavelet distributed in space at $t=0$:
\begin{equation}
\label{eq:3}
R(x)=\left(2\left(\pi f_{0}\left(x-1/f_{0}\right)\right)^{2} - 1\right) \cdot \exp\left(-\pi^2 f_{0}^2\left(x-1/f_{0}\right)^{2}\right),
\end{equation}
where $f_0$ is the central frequency. Due to the presence of time derivative terms in the viscous term, different viscoelastic wave equations result in different time-marching formulas, thereby significantly increasing the complexity of the problem. For simplicity, this paper focuses solely on viscoelastic wave equation of the form given by Eq. (\ref{eq:2}), assuming that the known right-hand side includes the $u_t$ term, but the viscous factor $\eta$ is unknown.

\subsection{Sparse Discovery}\label{sec:sparse-discovery}
Here, we review the fundamentals of the partial differential equation functional identification of nonlinear dynamics (PDE-FIND) algorithm. PDE-FIND is similar to SINDy but with the library including partial derivatives.

Let us consider the observations $u_m\in\mathbb{R}^{n_t^\prime\times n_x^\prime}$ over $n_t^\prime$ time points and $n_x^\prime$ spatial locations on a coarse grid. Upon flattening $u_m$ into a column vector $\mathbf{U}\in\mathbb{R}^{n_t^\prime\cdot n_x^\prime\times1}$, it is possible to establish a library $\boldsymbol{\Theta}\left(\mathbf{U}\right)\in\mathbb{R}^{n_t^\prime\cdot n_x^\prime\times D}$ consisting of $D$ candidate functional terms,
\begin{equation}
\label{eq:4}
\boldsymbol{\Theta}(\mathbf{U})=\left[u, u^{2}, \cdots, \nabla_{x} u, \nabla_{x}^{2} u, \nabla_{x} u \cdot u, \cdots, u_{t}, \nabla_{x} u \cdot u_{t} \cdots\right] .
\end{equation}

\begin{figure*}
	\centering
	\includegraphics[width=.85\textwidth]{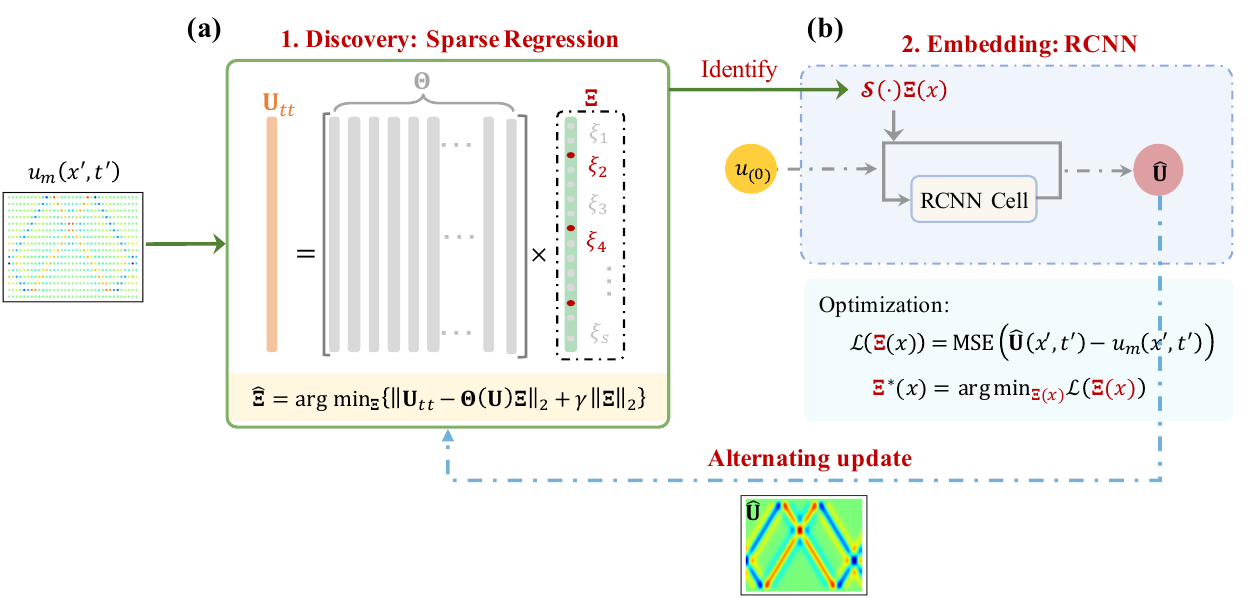}
	\caption{Schematic diagram of the proposed framework. (a) Discovery phase preliminarily identifies governing equations from observations. (b) Embedding phase use the RCNN model further optimizes the the imperfect function terms $\boldsymbol{\mathcal{S}}(\cdot)$ and coefficients $\boldsymbol{\Xi}(x)$ identified from discovery pahse. Given the measurements $u_m\left(x^\prime,t^\prime\right)\in\mathbb{R}^{n_t^\prime\times n_x^\prime}$ on a coarse grid (with resolution $n_t^\prime\times n_x^\prime)$, RCNN model can provide the discovered equation, inverted coefficients ($c$ and $\eta$), and high-resolution wavefield prediction $\widehat{\mathbf{{U}}} \in \mathbb{R}^{n_t\times n_x}$. The alternating update strategy implies that the high-resolution wavefield predicted by RCNN are provided to the discovery phase.}

        \label{fig:1}
\end{figure*}

Each column of the library $\boldsymbol{\Theta}\left(\mathbf{U}\right)$ corresponds to a particular candidate term of the governing equation, as shown in Fig. \ref{fig:1}. The evolution of the PDE can be represented in this library as follows:

\begin{equation}
\label{eq:5}
\mathbf{U}_{t t}=\boldsymbol{\Theta}(\mathbf{U}) \boldsymbol{\Xi},
\end{equation}
where $\boldsymbol{\Xi}=\{\xi_1,\xi_2,\cdots,\xi_D\} \in \mathbb{R}^{D\times1}$ is the coefficient vector, and each non-zero term in $\boldsymbol{\Xi}$ corresponds to a term in the PDE, e.g., $c$ and $\eta$ in Eq. (\ref{eq:2}). The requirement of sparse discovery is that each coefficient in $\boldsymbol{\Xi}$ is a scalar value.

However, due to the heterogeneity of the medium, the coefficients in $\boldsymbol{\Xi}$ should be spatially varying, i.e., $\boldsymbol{\Xi}\left(x\right)=\{\boldsymbol{\xi}_1(x),\boldsymbol{\xi}_2(x),\cdots,\boldsymbol{\xi}_D(x)\}$, which will be further optimized in the embedding phase. Specifically, sparse regression provides reasonable functional terms and scalar initial coefficients for further optimization of the subsequent RCNN model. The schematic diagram of our proposed framework is illustrated in Fig. \ref{fig:1}. Theoretical support for the RCNN model in embedding phase can be found in Section \ref{sec:recurr-conv-neur}.

We assume that a sufficiently complete function library implies that all functional forms of the representation are fully contained in it. Given a library of candidate function $\boldsymbol{\Theta}\left(\mathbf{U}\right)$, sparse regression aims to find a suitable coefficient vector such that it satisfies the sparsity requirement and has a small regression error. To effectively discover the physical laws of a system, the resulting model must be easy to interpret. This means that the solution of library $\boldsymbol{\Theta}\left(\mathbf{U}\right)$ should contain only a finite number of terms in the final discovered model. To do this, various techniques and methods can be utilized to promote sparsity of the coefficient vector $\boldsymbol{\Xi}$, such as least absolute shrinkage and selection operator (LASSO) \cite{Tibshirani_1996_RegressionShrinkageSelection}, sequential thresholded least-squares (STLS) \cite{Brunton_2016_DiscoveringGoverningEquations}, etc. In this study, it is achieved by using the sequential threshold ridge regression (STRidge) algorithm \cite{Rudy_2017_DatadrivenDiscoveryPartial} to solve the optimization problem described as
\begin{equation}
\label{eq:6}
\widehat{\boldsymbol{\Xi}}=\arg\min _{\boldsymbol{\Xi}}\left\{\left\|\mathbf{U}_{t t}-\boldsymbol{\Theta}(\mathbf{U}) \boldsymbol{\Xi}\right\|_{2}+\gamma\|\boldsymbol{\Xi}\|_{2}\right\},
\end{equation}
where $\|\boldsymbol{\Xi}\|_{2}$ measures the sparsity of the coefficient vector, $\left\|\mathbf{U}_{t t}-\boldsymbol{\Theta}(\mathbf{U}) \boldsymbol{\Xi}\right\|_{2}$denotes the regression error, $\gamma$ is the coefficient that balances the sparsity and regression error.

\subsection{Recurrent convolutional neural network}\label{sec:recurr-conv-neur}
\subsubsection{Time-marching scheme in RNN}
Here, we first established the conceptual comparison of viscoelastic wave physics system and a standard RNN \cite{Hughes_2019_WavePhysicsAnalog}. The connections between RNN nodes form a directed graph along the temporal sequence, allowing RNN to model temporal dynamics. At a given time step \(t\), an RNN operates on the current input vector \(\boldsymbol{x}_{(t)}\) in the sequence and the previous hidden state vector \(\boldsymbol{h}_{(t-1)}\) to produce an output vector \(\boldsymbol{y}_{(t)}\) and an updated hidden state \(\boldsymbol{h}_{(t)}\), which can be described by the following update equation,
\begin{equation}
\label{eq:7}
\begin{aligned}
\boldsymbol{h}_{(t)} &=\sigma^{(h)}\left(\boldsymbol{W}_{h h} \cdot \boldsymbol{h}_{(t-1)}+\boldsymbol{W}_{x h} \cdot \boldsymbol{x}_{(t)}\right), \\
\boldsymbol{y}_{(t)} &=\sigma^{(y)}\left(\boldsymbol{W}_{yh} \cdot \boldsymbol{h}_{(t)}\right),
\end{aligned}
\end{equation}
where $\boldsymbol{W}_{hh}$, $\boldsymbol{W}_{xh}$, and $\boldsymbol{W}_{yh}$ are dense matrices optimized during training. $\sigma^{(h)}$ and $\sigma^{(y)}$ are nonlinear activation functions. The hidden state $\boldsymbol{h}_{t}$ at the current time step $t$ will be involved in computing the hidden state $\boldsymbol{h}_{t+1}$ at the next time step $t+1$. In the next time step, the definition of the hidden state is the same as that used in the previous time step, so the calculation in Eq. (\ref{eq:7}) is recurrent.

\begin{figure*}
	\centering
	\includegraphics[width=.8\textwidth]{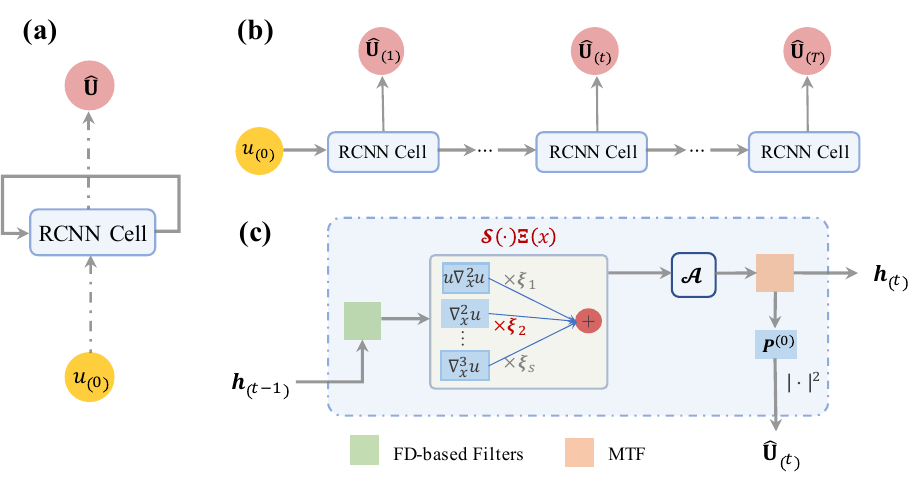}
	\caption{The architecture of RCNN model in embedding phase. (a) The directed acycic graph of a RCNN for the forward modelling. (b) The unrolled directed acyclic graph of the RCNN. $\widehat{\mathbf{U}}_{(1)}$ and $\widehat{\mathbf{U}}_{(T)}$ are the predicted wavefield of the RCNN cell at times $t_{1}$ and $t_{T}$, respectively. (c) The single RCNN cell architecture. $\boldsymbol{\mathcal{S}}(\cdot)\boldsymbol{\Xi}(x)$ is provided by the sparse regression from the previous discovery phase.}
        \label{fig:2}
\end{figure*}

In each alternating update of the discovery-embedding cycle, we define the equation obtained using sparse regression in the discovery phase as,
\begin{equation}
\label{eq:8}
u_{tt} = \boldsymbol{\mathcal{S}}\left(u\right)\boldsymbol{\Xi}(x) + \hat{\eta} u_{t},
\end{equation}
where $\boldsymbol{\mathcal{S}}(\cdot)$ contains the spatially correlated PDE operators discovered by sparse regression, $\boldsymbol{\Xi}(x)$ is the corresponding coefficient matrix discovered. $\boldsymbol{\mathcal{S}}\left(u\right)\boldsymbol{\Xi}(x)$ represents one or more terms in the PDE obtained by multiplying function terms with corresponding coefficients and accumulating them. Note that $\hat{\eta}$ is the coefficient in $\boldsymbol{\Xi}(x)$ corresponding to $u_t$ in the discovery phase. For simplicity, we extract $\hat{\eta}$ from $\boldsymbol{\Xi}(x)$ and express it explicitly here.

Using a time increment of $\delta t$ to discretize Eq. (\ref{eq:8}) with finite difference scheme, we obtain the recursive relation
\begin{equation}
\label{eq:9}
\frac{u_{(t+1)}-2 u_{(t)}+u_{(t-1)}}{\delta t^{2}} = \boldsymbol{\mathcal{S}}\left(u_{(t)}\right)\boldsymbol{\Xi}(x) + \hat{\eta} \frac{u_{(t+1)}-u_{(t-1)}}{2\delta t}.
\end{equation}

The above equation can be written in matrix form
\begin{equation}
\label{eq:10}
\left[\begin{array}{c}
u_{(t+1)} \\
u_{(t)}
\end{array}\right]
=\left[\begin{array}{cc}
\frac{4+2 \delta t^{2} \boldsymbol{\mathcal{S}}(\cdot)\boldsymbol{\Xi}(x)}{2-\hat{\eta} \delta t} & -\frac{2+\hat{\eta} \delta t}{2-\hat{\eta} \delta t} \\
1 & 0
\end{array}\right] \cdot\left[\begin{array}{c}
u_{(t)} \\
u_{(t-1)}
\end{array}\right],
\end{equation}
where the middle matrix may made up of matrices.

To relate Eq. (\ref{eq:10}) to the RNN update equation in Eq. (\ref{eq:7}), we define
\begin{equation}
\label{eq:11}
\boldsymbol{\mathcal{A}} = \left[\begin{array}{cc}
\frac{4+2 \delta t^{2} \boldsymbol{\mathcal{S}}(\cdot)\boldsymbol{\Xi}(x)}{2-\hat{\eta} \delta t} & -\frac{2+\hat{\eta} \delta t}{2-\hat{\eta} \delta t} \\
1 & 0
\end{array}\right].
\end{equation}

The hidden state $\boldsymbol{h}_t$ of the wave system is defined as the concatenation of the wavefields at time steps $t$ and $t+1$,
\(\boldsymbol{h}_t \equiv [u_{(t+1)}, u_{(t)}]^T\). Then, the update equation for the discoverd wave system defined by Eq. (\ref{eq:8}) can be rewritten as
\begin{equation}
\label{eq:12}
\begin{array}{l}
\boldsymbol{h}_{(t)}=\boldsymbol{\mathcal{A}} \cdot \boldsymbol{h}_{(t-1)}, \\
\mathbf{U}_{(t)}=\left|\boldsymbol{P}^{(\mathrm{o})} \cdot \boldsymbol{h}_{(t)}\right|^{2},
\end{array}
\end{equation}
where $\mathbf{U}_{(t)}$ represents the measurement output at a fixed receiver position and time, corresponding to $\boldsymbol{y}_{t}$ in Eq. (\ref{eq:7}). The connection between hidden state $\boldsymbol{h}_{t}$ and outputs $\mathbf{U}_{(t)}$ in the wave RNN is defined by the linear operator \(\boldsymbol{P}^{(o)}\), playing a similar role to the weight matrix $\boldsymbol{W}_{yh}$ in the output layer of standard RNN. According to Hughes et al. \cite{Hughes_2019_WavePhysicsAnalog} definition, $\boldsymbol{P}^{(\mathrm{o})} \equiv\left[\boldsymbol{M}^{(\mathrm{o}) T}, \mathbf{o}\right]$, where the linear operator $\boldsymbol{M}^{(o)}$ defines the corresponding spatial distribution of measurement points, and $\boldsymbol{o}$ is a matrix of all zeros.

It is noteworthy that sparse regression identifies the wave equation without a source term from observations. Consequently, the wave RNN described by Eq. (\ref{eq:12}) effectively corresponds to a standard RNN devoid of an input vector and its associated weight matrix ($\boldsymbol{x}_{(t)}$ and $\boldsymbol{W}_{xh}$ in Eq. (\ref{eq:7})). If the source input is continuously applied at each time step, then $\boldsymbol{x}_{(t)}$ will represent the source time function, while $\boldsymbol{W}_{xh}$ will denote the spatial vector of the source injection location. In a standard RNN, $\boldsymbol{W}_{hh}$ captures contributions from the previous state. It can be trained to learn how to use the hidden variables from previous time steps in the current time step. In contrast, in wave RNN, we leverage physical priors from FDTD by fixing $\boldsymbol{\mathcal{A}}$ as an untrainable matrix described by Eq. (\ref{eq:11}).

So far, we have shown that the time-marching scheme of the viscoelastic wave equation can be directly mapped to the framework of an RNN. Fig. \ref{fig:2} shows the architecture of the RCNN model at three levels of granularity.

\subsubsection{Finite difference-based filters in CNN} \label{sec:finite-diff-based}
The efficacy of a classical numerical PDE solver relies on the translational similarity of its discretized local differential operators \cite{Wu_2023_NeuralPartialDifferential}. From a machine learning perspective, these “translational similar” differential operators resemble the concept of convolution operators that function as the cornerstone to embed the “translational invariant” priors into neural networks \cite{LeCun_1995_ConvolutionalNetworksImages}.

Previous work investigated a deep relationship between convolutions and differentiations and discussed the connection between the order of the sum rules of the filters and the order of the differentiation operators \cite{Cai_2012_ImageRestorationTotal, Dong_2017_ImageRestorationWavelet}. By utilizing the concept of "translation invariance" observed in PDE differential operators, we can approximate discretization schemes of finite difference operators using a series of convolution kernels in CLs.

\begin{figure*}
	\centering
	\includegraphics[width=.8\textwidth]{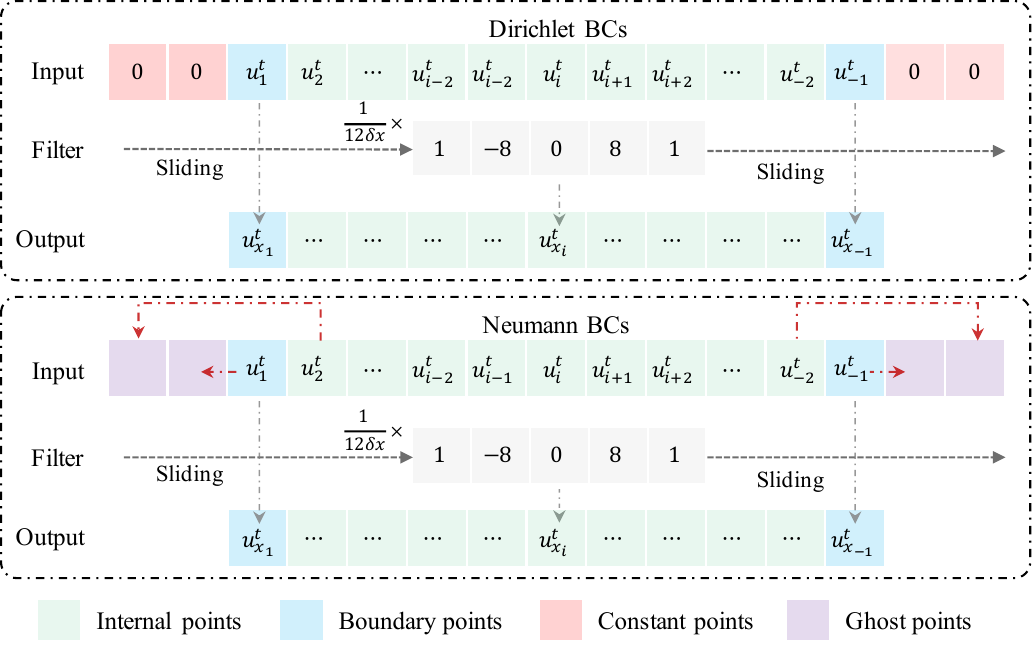}
	\caption{Schematic diagram of FD-based filters and hard embedding of boundary conditions. The filter $\mathcal{K}_{x}=\frac{1}{12\delta x}(1,-8,0,8,1)$ implements a 4th-order FD derivative operation in 1D CNN. For fixed boundary condition (upper figure), directly fill constant $0$ on the extended nodes outside the boundary. For free-surface boundary condition (lower figure), determine the filling values of ghost points based on the values of the internal wavefield.}
        \label{fig:3}
\end{figure*}

We apply second-order and fourth-order central difference schemes separately in the temporal and spatial dimensions. A uniquely determined gradient-free "frozen" filter is implemented on an unbiased convolution layer. The receptive field is defined as the input region that each FD-based filter focuses on. In the temporal direction, the filter $\mathcal{K}_{t}$ and receptive field $F(u_{x_{i}}^{t})$ for implementing the first-order derivative using a three-point difference scheme are defined as:
\begin{equation}
\label{eq:13}
\begin{aligned}
F(u_{x}^{t_{i}}) &=\left[u_{x}^{t_{i-1}}, u_{x}^{t_{i}}, u_{x}^{t_{i+1}}\right], \\
\mathcal{K}_{t} &= \left(-1,0,1\right) \times \frac{1}{2\delta t},
\end{aligned}
\end{equation}
where $\delta t$ is the time interval. The receptive field $F(u_{x_{i}}^{t})$ includes three adjacent temporal solution variables $u$. Then, the first-order time derivative of $u$ can be expressed in CNN as:
\begin{equation}
\label{eq:14}
\frac{\partial u_{x}^{t_{i}}}{\partial t} \approx \frac{1}{2\delta t}(-1,0,1)\left[u_{x}^{t_{i-1}}, u_{x}^{t_{i}}, u_{x}^{t_{i+1}}\right]= \mathcal{K}_{t}\cdot  F(u_{x}^{t_{i}}).
\end{equation}

Observing the above equation, it can be found that the dot product between the designed filter and the corresponding receptive field is equivalent to the FD derivative process.
Differential in spatial dimension are consistent with time, but require more neighboring points to be involved in the differentiation process.
Fig. \ref{fig:3} illustrates the derivative process of FD-based filters in 1D CLs using $\mathcal{K}_x$ as an example. The derivatives of each order in the spatial direction can be expressed as,
\begin{equation}
\label{eq:15}
\begin{aligned}
\frac{\partial u_{x_{i}}^{t}}{\partial x} &\approx \mathcal{K}_{x}\cdot  F(u_{x_{i}}^{t}) = \frac{1}{12\delta x}(1,-8,0,8,1) \cdot F(u_{x_{i}}^{t}), \\
\frac{\partial^2 u_{x_{i}}^{t}}{\partial x^2} &\approx \mathcal{K}_{xx}\cdot  F(u_{x_{i}}^{t}) = \frac{1}{12\delta x^2}(-1,16,-30,16,-1) \cdot F(u_{x_{i}}^{t}), \\
\frac{\partial^3 u_{x_{i}}^{t}}{\partial x^3} &\approx \mathcal{K}_{xxx}\cdot  F(u_{x_{i}}^{t}) = \frac{1}{12\delta x^3}(-1,2,0,-2,1) \cdot F(u_{x_{i}}^{t}), \\
\end{aligned}
\end{equation}
where $\delta x$ is the spatial mesh size. $\mathcal{K}_{x},\ \mathcal{K}_{xx}$ and $\mathcal{K}_{xxx}$ denote the first three orders of filters corresponding to the five-point center difference scheme, respectively. The receptive field is composed of the solution variables from five spatially adjacent positions, denoted as $F(u_{x_{i}}^{t})=\left[u_{x_{i-2}}^{t}, u_{x_{i-1}}^{t}, u_{x_{i}}^{t}, u_{x_{i+1}}^{t}, u_{x_{i+2}}^{t} \right]$.

\subsubsection{Optimization}\label{sec:optimization}
Given the measurements $u_m\left(x^\prime,t^\prime\right)\in\mathbb{R}^{n_t^\prime\times n_x^\prime}$ on a coarse grid (with resolution $n_t^\prime\times n_x^\prime)$, the embedding phase is optimized by minimizing the mean square error (MSE) between model predictions and coarse measurements. Specifically, the loss function of the RCNN model is defined as:
\begin{equation}
\label{eq:16}
\mathcal{L}(\boldsymbol{\Xi}(x))=\operatorname{MSE}\left(\hat{\mathbf{U}}\left(x^{\prime}, t^{\prime}\right)-u_{m}\left(x^{\prime}, t^{\prime}\right)\right),
\end{equation}
where $\hat{\mathbf{U}}\left(x^\prime,t^\prime\right)$ denotes the mapping of the high-resolution prediction $\hat{\mathbf{U}}$ on the coarse grid $\left(x^\prime,t^\prime\right)$.

The training process of the parameterized RCNN is equivalent to searching for the optimal parameters in a function space, in order to learn governing equation of the dynamical system and achieve high-resolution prediction from observational data. Utilizing the coefficients corresponding to predetermined function terms from sparse regression as initial estimates, these trainable parameters are then optimized in the RCNN model:
\begin{equation}
\label{eq:17}
\boldsymbol{\Xi}^*(x) = \arg\min_{\boldsymbol{\Xi}(x)} \left\{ \mathcal{L}(\boldsymbol{\Xi}(x)) \right\}.
\end{equation}

We optimize the RCNN model using an Adam optimizer with a stochastic gradient descent method \cite{Kingma_2014_AdamMethodStochastic} followed by optimization with the L-BFGS optimizer \cite{Liu_1989_LimitedMemoryBFGS}. Training with the L-BFGS optimizer stops when the maximum number of iterations is reached or convergence is achieved. We mainly use the Adam optimizer to provide a good starting point for L-BFGS and to speed up the training process. We add a filtering mechanism to the RNN cell such that if the average value of any term in $\boldsymbol{\Xi}(x)$ is less than $10^{-3}$ during the optimization process, the corresponding term will be removed from the discovered equation.

Furthermore, we introduce a alternating update strategy that effectively enhances recognition efficiency and accuracy for the case of homogeneous media. Specifically, we first utilize sparse regression to identify the initial equation from sparse and noisy data $u_m\left(x^\prime,t^\prime\right)$. Then employ Adam and L-BFGS optimizers in RCNN model to provide clean high-resolution results $\hat{\mathbf{U}}\left(x,t\right)$ for the next cycle of sparse regression. In our framework, the initial sparse regression employs downsampled noisy data to compute derivatives. The subsequent sparse regression in the next cycle leverages high-resolution clean data obtained from the previous RCNN model to compute the derivatives. Both sparse regression and RCNN models are implemented using the finite difference-based filter described in Section \ref{sec:finite-diff-based}.

\subsection{Boundary conditions setting}\label{sec:bound-cond-sett}
The boundary condition $u=0$ in a wave equation reflects the wave, but $u$ changes sign at the boundary, while the condition $u_x=0$ reflects the wave as a mirror and preserves the sign, called the homogeneous Dirichlet or Neumann conditions, respectively. For simulation problems involving the free surface, the implementation of the boundary condition is crucial, i.e., satisfying $\frac{\partial u}{\partial\mathbf{n}}\equiv\mathbf{n}\cdot\nabla u=0$, where the derivative $\partial/\partial \mathbf{n}$ is in the outward normal direction from a general boundary. For a 1D domain $[0, L]$, we have that $\left.\frac{\partial u}{\partial\mathbf{n}}\right|_{x=L}=\frac{\partial u}{\partial x}=0$.

For homogeneous Dirichlet BCs, it can be strictly incorporated into the solution variables through time-invariant padding operation. For homogeneous Neumann BCs, we extended the mesh using ghost points \cite{Wang_2019_FourthOrderFinite}, and derived the padding values of the ghost points based on the 1D free-surface boundary condition $u_x=0$. In this way, applying the standard difference scheme at the boundary point will be correct and will ensure that the solution is compatible with the boundary conditions. Fig. \ref{fig:3} shows the details of applying homogeneous Dirichlet and Neumann boundary conditions in 1D CNN.

\begin{figure*}
	\centering
	\includegraphics[width=.8\textwidth]{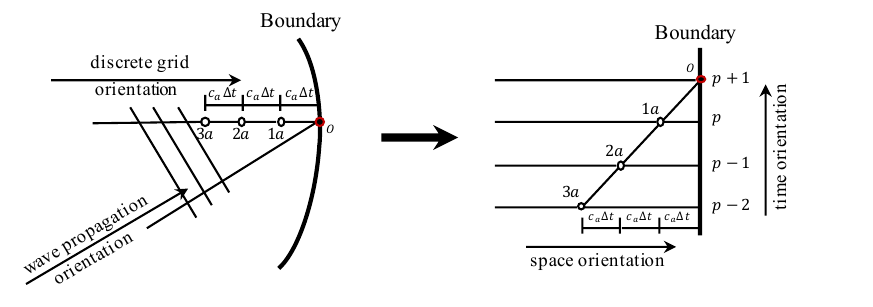}
	\caption{Schematic diagram of multi-transmitting formula (MTF).}
        \label{fig:4}
\end{figure*}

We consider the problem of identifying and inverting the wave equation in energy-closed systems versus energy-open systems with simulated real subsurface half-space. For wave simulation of energy-open system, absorbing boundary conditions (ABCs) are typically needed to prevent spurious wave reflections at the boundaries. In this paper, we employ the multi-transmitting formula (MTF) to satisfy the transmission of outgoing waves at the bottom boundary \cite{Liao_1984_TransmittingBoundaryNumerical, Xing_2021_TheoryNewUnified, Xing_2021_SpectralelementFormulationMultitransmitting}. In MTF, the motion of an arbitrary artificial boundary node at each timestep is directly predicted from the motions of some adjacent nodes at several previous timesteps. It is expressed by a discrete formula as
\begin{equation}
\label{eq:18}
u_{0}^{p+1}=\sum_{j=1}^{N}(-1)^{j+1} C_{j}^{N} u_{j a}^{p+1-j},
\end{equation}
where $N$ is the transmitting order and $C_j^N=N!/(j!\left(N-j\right)!)$ are binomial coefficients. $u_0^{p+1}=u(0,\left(p+1\right)\delta t)$ denotes the motion of artificial boundary node 0 at time instant $\left(p+1\right)\delta t$, as shown in Fig. \ref{fig:4}. $u_{ja}^{p+1-j}=u(jc_a\delta t,\left(p+1-j\right)\delta t)$ denotes the motion of a uniformly distributed node $ja$ on a discrete grid line pointing from point 0 to the inner domain at the moment $\left(p+1-j\right)\delta t$. $c_a$ is a user-defined hyperparameter called artificial wave velocity.

The formula of MTF yields to $u_0^{p+1}=u_{1a}^p$ for $N=1$, $u_0^{p+1}=2u_{1a}^p-u_{2a}^{p-1}$ for $N=2$, and $u_0^{p+1}=3u_{1a}^p-3u_{2a}^{p-1}+u_{3a}^{p-2}$ for $N=3$. The MTF is combined with the inner-domain discrete format by interpolating the motion of each MTF computation point $1a,2a,\cdots,Na$ with the motion of nearby inner-domain nodes. This interpolation is easy to implement and offers multiple selectable options. In this work, we adopt the 2nd order MTF ($N=2$) numerical scheme with quadratic interpolation \cite{Liao_1992_NumericalInstabilitiesLocal}.

\section{Results}\label{sec:numer-exper}
In this section, we validate the performance of our proposed method with some numerical results of potential practical significance, considering wave propagation in elastic/viscoelastic, homogeneous/inhomogeneous media.

We utilize the finite difference method (FDM) to generate datasets, employing both second-order central difference scheme and fourth-order central difference scheme for temporal and spatial discretization. The number of grid points in the discrete spatial domain is 128, and the number of discrete points in the temporal domain is determined by the Courant–Friedrichs–Lewy (CFL) condition. To ensure the stability of the finite difference method, we set the time interval to $\delta t=0.9\delta x/\sqrt c$.

As mentioned in Section \ref{sec:problem-statement}, we assume that the viscosity is controlled by the $u_t$ term in all cases of this work. Moreover, during the sparse regression phase, the viscous term will be exempted from filtering. In the first identification, $\eta$ is set to 0.1 as the initial value for inversion. Pareto analysis is used to choose the appropriate $\gamma$ in Eq. (\ref{eq:6}) to balance model complexity and regression accuracy \cite{Rudy_2017_DatadrivenDiscoveryPartial, Rao_2023_EncodingPhysicsLearn}. The learning rate of the Adam optimizer of RCNN model is initialized to $5 \times 10^{-3}$.

To quantitatively evaluate the performance of the proposed method, two types of relative $\mathcal{L}_2$ errors are defined in terms of the wavefield and wave velocity. Across all cases, the relative $\mathcal{L}_2$ error of the wavefield, defined as $\epsilon(\mathbf{U})=\left\|\widehat{\mathbf{U}}-\mathbf{U}_{\text {true }}\right\|_{2} /\left\|\mathbf{U}_{\text {true }}\right\|_2$, measures the relative distance between the predicted wavefield $\widehat{\mathbf{U}}$ and the ground truth $\mathbf{U}_{\text{true}}$. On the other hand, the relative $\mathcal{L}_2$ error of the wave velocity $\epsilon(c)=\left\|\hat{c}-c_{\text {true }}\right\|_{2} /\left\|c_{\text {true }}\right\|_2$ represents the difference between the inverted wave velocity $\hat{c}$ and the exact wave velocity $c_{\text{true}}$ in heterogeneous media.

\subsection{Wave equation in an elastic homogeneous medium}
We first consider two cases of wave propagation in elastic homogeneous media. In Case 1, the reflection phenomenon of wave propagation in a energy-closed system is modeled by imposing homogeneous Dirichlet boundary conditions at both ends of the spatial medium. In Case 2, we impose a homogeneous Neumann boundary condition $u_x=0$ at the top of the medium ($x=L$) and MTF at the bottom of the medium to simulate wave propagation in a 1D half-space. The finite difference method is used to solve Eq. (\ref{eq:1})  and generate synthetic data. In this case, the spatial range is $x\in\left[0,\ 6\right]$, the temporal range is $t\in\left[0,\ 5\right]$, the wave velocity $c$ is $2.5$. The number of grid points in the spatial domain is 128 and the time domain is discretized into 420 uniform points. The source is represented by the initial distribution of the Ricker wavelet on $x$, with a central frequency of $f_0=0.5$.

We performed downsampling on the wavefield in both spatial and temporal dimensions, with downsampling factors of 8 and 12 respectively. This results in a measurment resolution of only $16 \times 35$ provided to the sparse regression algorithm, accounting for $1.04\%$ of the synthetic dataset. The candidate functions in the function library are designed to capture the different forms of wave equations that may appear in the measurement. We employ a dictionary denoted as $\boldsymbol{\Theta}\left(\mathbf{U}\right)\in\mathbb{R}^{35\cdot16\times60}$, which contains 60 candidate functions. These candidate functions are constructed to include polynomial terms up to the third order $\left\{1,u,u^2,u^3\right\}$ and derivatives up to the third order $\left\{1,u_x,u_x^2,u_{xx},u_{xx}^2,u_{xxx},u_{xxx}^2,u_t\right\}$, as well as combinations of them.

\begin{figure}
	\centering
	\includegraphics[width=.4\textwidth]{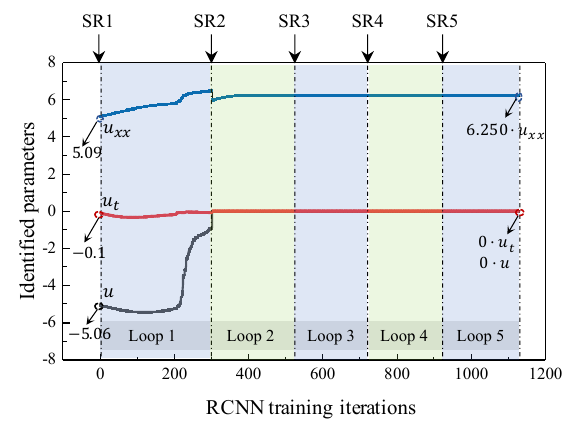}
	\caption{Case 1: Discovery-embedding alternately updates the identified function terms and corresponding coefficients. In the first execution of sparse regression (SR1), the discovered equation is $u_{tt} = 5.09 u_{xx} - 5.06 u - 0.1 u_t$. After the fifth loop (Loop 5), the correct equation $u_{tt}=6.25 u_{xx}$ is obtained.}
        \label{fig:5}
      \end{figure}

\begin{figure}
	\centering
	\includegraphics[width=.4\textwidth]{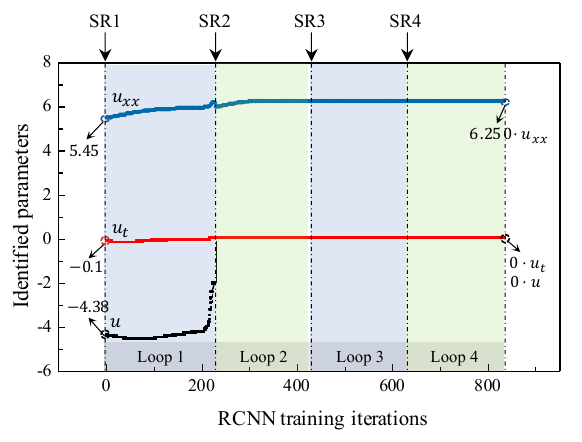}
       	\caption{Case 2: Discovery-embedding alternately updates the identified function terms and corresponding coefficients. In the first execution of sparse regression (SR1), the discovered equation is $u_{tt} = 5.45 u_{xx} - 4.38 u - 0.1 u_t$. The RCNN model in the fourth loop (Loop 4) outputs the correct equation $u_{tt}=6.25 u_{xx}$. }
        \label{fig:6}
\end{figure}

\begin{figure*}
	\centering
	\includegraphics[width=.8\textwidth]{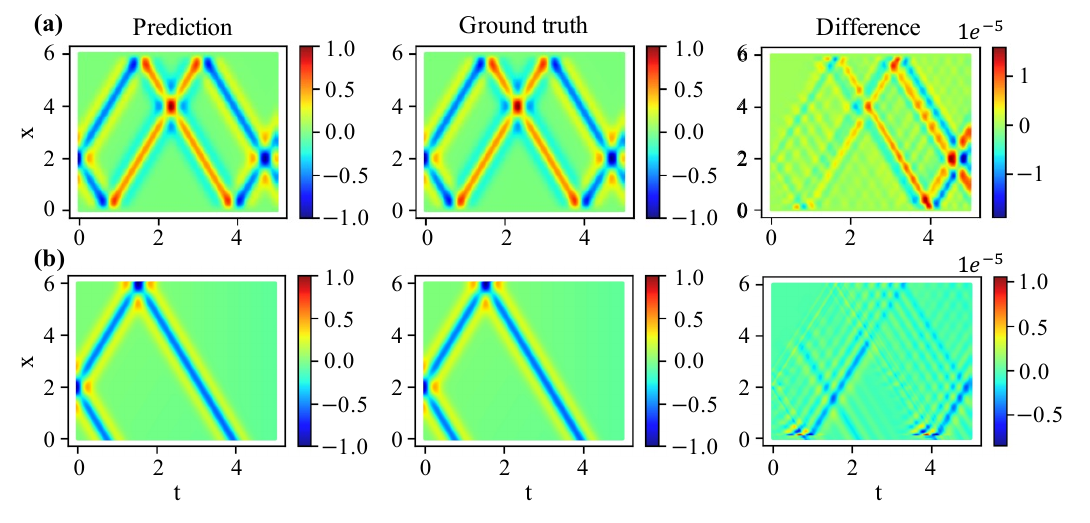}
	\caption{Comparison between predicted wavefield and ground truth. (a) Case 1. (b) Case 2.}
        \label{fig:7}
\end{figure*}

Due to the extremely sparse wavefield measurement, the initial sparse regression provides inaccurate results for both two cases, as shown in Figs. \ref{fig:5} and \ref{fig:6}. For Case 1, the identified equation after initial sparse regression is $u_{tt}=5.09u_{xx}-5.06u-0.1u_t$, and $u_{tt}=5.45u_{xx}-4.38u-0.1u_t$ for Case 2. However, in the subsequent RCNN optimization, the coefficients corresponding to the erroneous terms $u$ and $u_t$ gradually approach zero. We executed a total of 5 loops for Case 1 and 4 loops for Case 2. In each of loop we first performed sparse regression, followed by 200 epochs of training using the Adam optimizer. The L-BFGS was used by up to 100 epochs of optimization, as the L-BFGS optimizer may stop early. From the sparse regression of the second loop, the correct function terms are obtained. In the following iterations, the coefficients are fine-tuned, resulting in highly accurate equations. Once the training is completed, a high-resolution solution can be inferred from the trained model, as shown in Fig. \ref{fig:7}.

\subsection{Case 3: viscoelastic wave equation in a homogeneous medium}
\begin{figure}
	\centering
	\includegraphics[width=.4\textwidth]{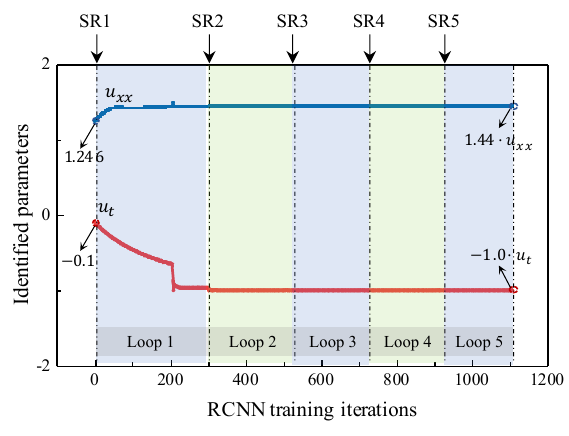}
       	\caption{Case 3: Discovery-embedding alternately updates the identified function terms and corresponding coefficients. In the first execution of sparse regression (SR1), the discovered equation is $u_{tt}=1.246u_{xx}-0.1u_t$. After the fifth loop (Loop 5), the correct equation $u_{tt}=1.44 u_{xx}-1.0u_t$ is obtained.}
        \label{fig:8}
\end{figure}

\begin{figure*}
	\centering
	\includegraphics[width=.8\textwidth]{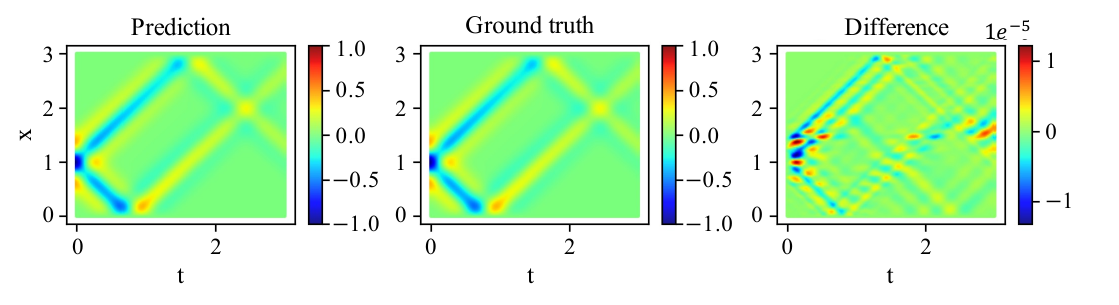}
	\caption{Comparison between the predicted wavefield and ground truth in Case 3.}
        \label{fig:9}
\end{figure*}

In the development of the Cases 1 and 2, a perfectly elastic medium was considered. For studies where small deformations occur, many elastic materials do not deviate grossly from perfectly elastic behavior. It is well-known, however, that when materials are set in vibration, the vibrations are accompanied by dissipation, due to the conversions of elastic energy to internal energy \cite{Charlier_1986_WaveEquationsLinear}. In Case 3, we consider the wave equation in a viscoelastic, homogeneous and energy-closed system. In this case, the spatial range is $x\in\left[0,\ 3\right]$, the temporal range is $t\in\left[0,\ 3\right]$, and we set $c=1.2$, $\eta=-1.0$. The number of grid points in the spatial domain is 128 and the time domain is discretized into 242 uniform points. The source is represented by the initial distribution of the Ricker wavelet on $x$, with a central frequency of $f_0=1.0$.

We downsample the wavefield in the spatial and temporal dimensions with reduction factors of 8 and 12, respectively. This results in a wavefield data size of only $16\times21$, which accounts for a mere $1.08\%$ of the synthetic dataset. In alignment with Case 1 and Case 2, we construct a dictionary representation denoted as $\boldsymbol{\Theta}\left(\mathbf{U}\right)\in\mathbb{R}^{21\cdot16\times60}$, which contains 60 candidate functions.

Fig. \ref{fig:8} illustrates the progressive evolution of identified coefficients during five loops, with each loop utilizing a fixed Adam setting of 200 epochs. The initial outcome yielded by the sparse regression technique is $u_{tt}=1.246u_{xx}-0.1u_t$, disclosing imprecise wave velocity and viscosity coefficients attributed to the limited number of measurements available. However, significantly improved outcomes are achieved following the RCNN optimization in the first loop. Moving on to Fig. \ref{fig:9}, a comparative analysis is presented, where the predicted wavefield generated by our method is contrasted against results obtained using finite difference method, accompanied by their respective errors. Evidently, our method adeptly captures the phenomenon of wave attenuation in viscoelastic media.

\subsection{Case 4: viscoelastic wave equation with linearly varying wave velocity}
\begin{figure}
	\centering
	\includegraphics[width=.35\textwidth]{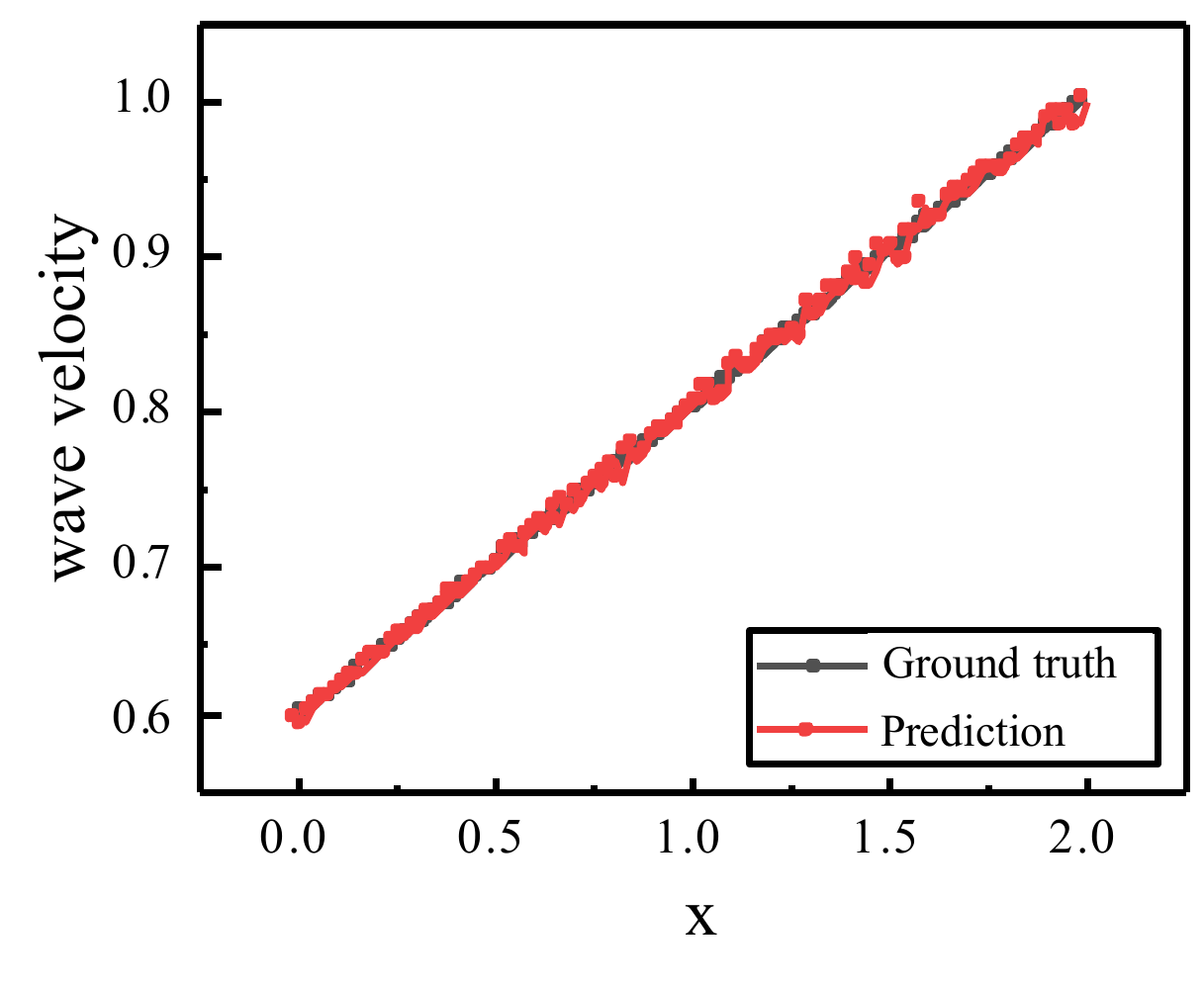}
	\caption{Comparison between the inverted wave velocity model and the ground truth.}
        \label{fig:10}
\end{figure}

\begin{figure*}
	\centering
	\includegraphics[width=.8\textwidth]{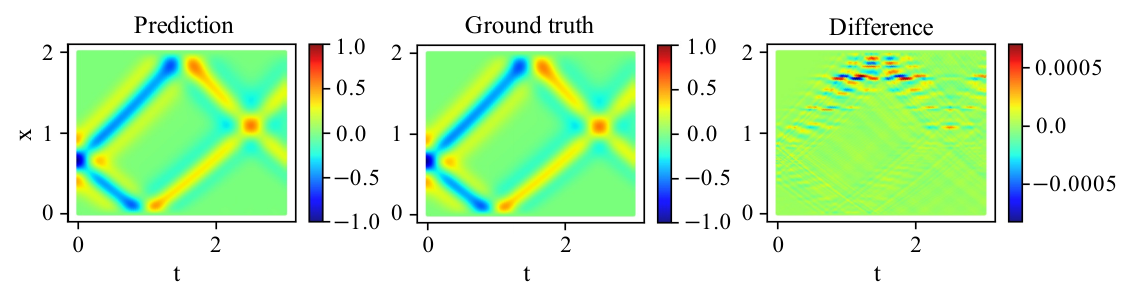}
       	\caption{Comparison between the predicted wavefield and ground truth in Case 4.}
        \label{fig:11}
\end{figure*}

In this subsection, we illustrate the performance of the proposed method in terms of the viscoelastic wave equation with linearly varying wave velocity. Homogeneous Dirichlet boundary conditions are imposed at both ends of the spatial medium and wave energy attenuation is considered simultaneously. The model is specifically set up with $x\in\left[0,\ 2\right]$, $t\in\left[0,\ 3\right]$, $\eta=-0.5$, and the wave velocity was set to vary linearly from 0.6 to 1.0. The central frequency $f_0$ of Ricker source is 1.5. The number of grid points in the spatial domain is 128 and the time domain is discretized into 302 uniform points.

The observations provided to sparse regression and RCNN training were downsampled to a coarse grid of $26 \times 31$, which represents only $2.09\%$ of the synthesized dataset. The candidate function library is the same as in the previous case, i.e., $\boldsymbol{\Theta}\left(\mathbf{U}\right)\in\mathbb{R}^{31\cdot26\times60}$. Although RCNN phase provides high-resolution measurements, the inherent limitations of sparse regression result in only being able to identify scalar wave velocity in each cycle. Therefore, the cycle training strategy mentioned in Section \ref{sec:optimization} was not employed in Case 4. At the beginning, sparse regression identified a scalar wave velocity of 0.739, which was used as an initial value for the RCNN optimization. We performed optimization on RCNN by employing the Adam optimizer for a total of 1000 epochs. Following that, we proceeded with additional optimization using L-BFGS for a maximum of 500 epochs.

The comparison between the predicted wave velocities and the true wave velocities is shown in Fig. \ref{fig:10}. It can be observed that the predicted wave velocities of the model closely match the true values. Furthermore, our method accurately identified the viscous factor, $\eta=-0.499926$, directly from the observed data. The optimized model is capable of directly predicting high-resolution ($128\times302$) wavefield. As shown in Figure \ref{fig:11}, the predicted wavefield accurately captures variations in wave velocity and energy dissipation, exhibiting small point-wise errors when compared to the ground truth.

\subsection{Ablation study}
Next, we consider a more practical scenario where the data is disturbed by noise. For this purpose, the previously generated synthetic data was added with up to $20\%$ noise. All the numerical implementations in this paper are coded in Pytorch \cite{Paszke_2019_PyTorchImperativeStylea} and performed on an NVIDIA GeForce RTX 3090 Card (24 GB). We statistically calculated the average computational cost for each test case for different noise levels, as shown in Table \ref{tab:1}. The identified RHS at different noise levels as well as the relative $\mathcal{L}_2$ errors of the wavefield $(\epsilon\left(\mathbf{U}\right))$ in Cases 1-3 are shown in Table \ref{tab:2}. For Case 4, the relative $\mathcal{l}_2$ errors of the predicted wavefield $(\epsilon\left(\mathbf{U}\right))$ and the inverted wave velocity $\epsilon(c)$, as well as identified $\eta$ by the proposed framework at different noise levels $\eta$ are shown in Table \ref{tab:3}.

It can be observed that the increase in noise level leads to inaccurate wave velocity identification and an increase in $\epsilon\left(\mathbf{U}\right)$. Nevertheless, the proposed method is able to successfully identify viscoelastic wave equation and velocity model even in the presence of a large amount of noise.

\begin{table}[width=.9\linewidth,cols=4,pos=h]
\caption{The total computation cost for each test case.}
\label{tab:1}
\begin{tabular*}{\tblwidth}{@{} LLLL@{} }
\toprule
Case 1 & Case 2 & Case 3 & Case 4\\
\midrule
14.2 min & 11.5 min & 13.8 min & 14.9 min \\
\bottomrule
\end{tabular*}
\end{table}

\begin{table*}[width=1.78\linewidth,cols=8,pos=h]
  \caption{Performance of the proposed method in the case of homogeneous media.}
  \label{tab:2}
  \centering
  \renewcommand\arraystretch{1.0}
\begin{tabular}{ccm{1.6cm}<{\centering}m{1.6cm}<{\centering}m{1.6cm}<{\centering}m{1.6cm}<{\centering}m{1.6cm}<{\centering}m{1.6cm}<{\centering}}
  \toprule
  \multirow{2}{*}{Cases} & \multirow{2}{*}{Metrics} & \multicolumn{6}{c}{Noise Level} \\
\cmidrule { 3 - 8 } & & \( 0 \% \) & \( 2 \% \) & \( 5 \% \) & \( 10 \% \) & \( 15 \% \) & \( 20 \% \) \\
  \midrule
  \multirow{2}{*}{Case 1} & RHS & \( 6.25019 u_{xx} \) & \( 6.24891 u_{xx} \) & \( 6.24895 u_{xx}\) & \( 6.24202 u_{xx} \) & \( 6.23343 u_{xx} \) & \( 6.22114u_{xx} \) \\
  \specialrule{0em}{2pt}{2pt}

                         & \(\epsilon\left(\mathbf{U}\right)\) & \( 2.509 \times 10^{-5} \) & \( 2.846 \times 10^{-3} \) & \( 1.793 \times 10^{-3}\) & \( 1.921 \times 10^{-2} \) & \( 2.794 \times 10^{-2} \) & \( 3.711 \times 10^{-2} \) \\
  \specialrule{0em}{2pt}{2pt}
  \multirow{2}{*}{Case 2} & RHS & \( 6.24989 u_{xx} \) & \( 6.251659 u_{xx} \) & \( 6.25494 u_{xx}\) & \( 6.26064 u_{xx} \) & \( 6.26668 u_{xx} \) & \( 6.3061 u_{xx} \) \\
      \specialrule{0em}{2pt}{2pt}

  & \(\epsilon\left(\mathbf{U}\right)\) & \( 5.178 \times 10^{-6} \) & \( 3.652 \times 10^{-3} \) & \( 9.652 \times 10^{-3}\) & \( 1.7724 \times 10^{-2} \) & \( 2.5986 \times 10^{-2} \) & \( 4.166 \times 10^{-2} \) \\
      \specialrule{0em}{2pt}{2pt}

  \multirow{2}{*}{Case 3} & RHS & \( 1.44004 u_{xx} - 1.0001 u_t \) & \( 1.44252 u_{xx}-0.9729 u_t \) & \( 1.43774 u_{xx}-0.9738u_t\) & \( 1.43784 u_{xx}-0.9678u_t \) & \( 1.36802 u_{xx}-0.9488u_t \) & \( 1.34835 u_{xx}-0.9226u_t \) \\
      \specialrule{0em}{2pt}{2pt}

 & \(\epsilon\left(\mathbf{U}\right)\) & \( 1.603 \times 10^{-5} \) & \( 1.832 \times 10^{-3} \) & \( 9.289 \times 10^{-2}\) & \( 4.723 \times 10^{-2} \) & \( 8.228 \times 10^{-2} \) & \( 5.514 \times 10^{-2} \) \\

    \bottomrule \\
\end{tabular}
\end{table*}

\begin{table*}[width=1.7\linewidth,cols=8,pos=h]
  \caption{Performance of the proposed method in the case of inhomogeneous media.}
  \label{tab:3}
  \centering
\begin{tabular}{@{} ccm{1.6cm}<{\centering}m{1.6cm}<{\centering}m{1.6cm}<{\centering}m{1.6cm}<{\centering}m{1.6cm}<{\centering}m{1.6cm}<{\centering} @{}}
  \toprule
  \multirow{2}{*}{Case} & \multirow{2}{*}{Metrics} & \multicolumn{6}{c}{Noise Level} \\
\cmidrule { 3 - 8 } & & \( 0 \% \) & \( 2 \% \) & \( 5 \% \) & \( 10 \% \) & \( 15 \% \) & \( 20 \% \) \\
  \midrule
  \multirow{3}{*}{Case 4} & \( \eta \) & \( -0.4999 \) & \( -0.4922 \) & \( -0.4926 \) & \( -0.4904 \) & \( -0.4801 \) & \( -0.4726 \) \\
  \specialrule{0em}{2pt}{2pt}
  & \( \epsilon(c) \) & \( 9.410 \times 10^{-3} \) & \( 2.431 \times 10^{-2} \) & \( 3.642 \times 10^{-2} \) & \( 3.918 \times 10^{-2} \) & \( 4.323 \times 10^{-2} \) & \( 3.166 \times 10^{-1} \) \\
 &   \( \epsilon\left(\mathbf{U}\right) \) & \( 2.418 \times 10^{-4} \) & \( 4.151 \times 10^{-3} \) & \( 8.869 \times 10^{-2} \) & \( 1.642 \times 10^{-2} \) & \( 2.476 \times 10^{-2} \) & \( 2.059 \times 10^{-1} \) \\

    \bottomrule \\
\end{tabular}
\end{table*}

\section{Discussion}\label{sec:discussion}
Physically meaningful PDEs exhibit dimensional homogeneity, which means that all terms in the expressions discovered from the data must have matching dimensional units. Dimensional analysis is a crucial component of the framework for discovering PDEs, as it facilitates the reduction of the number of variables and enables the formulation of expressions that reflect physical balance. Before engaging in symbolic regression, one can employ automated dimensional analysis \cite{Udrescu_2020_AIFeynmanPhysicsinspired} or utilize the Buckingham $\pi$ theorem \cite{Buckingham_1914_PhysicallySimilarSystems, Matchev_2022_AnalyticalModelingExoplanet} to perform multiplicative operations on variables and constants, rendering them dimensionless. Furthermore, recent work \cite{Tenachi_2023_DeepSymbolicRegression} has utilized deep reinforcement learning techniques to recover analytical symbolic expressions from physical data by learning unit constraints, thereby ensuring that the proposed solutions are dimensionally consistent. In the discovery phase, we employed a sparse regression, which resulted in the inability to derive the governing equations containing specific physical units. Prior to our discovery, we had already established specific dimensional physical priors, including the units related to model size and sensor data. Consequently, we can manipulate the magnitudes of the inputs $x$ and $u$ in the sparse regression framework, enabling us to infer the corresponding units of the coefficients (e.g., $c^2$). However, there is a lack of balance in physical units on both sides of the obtained equation. Taking Case 3 as an example, we found that the identified equation is $u_{tt}=1.44u_{xx}-u_t$, without incorporating the implicit relationship $c^2=1.44$. We acknowledge this limitation and are actively considering enhancements to the discovery phase to provide more robust and physically meaningful initial guesses for PDEs in future work.

The inherent limitations of sparse regression result in only being able to identify scalar wave velocity when alternating from RCNN to sparse regression. In the context of scalar-coefficient scenarios, the alternating optimization strategy introduced in this study enhances both the accuracy and robustness of sparse regression, particularly when dealing with observations that are characterized by sparsity and significant noise. In addressing the challenge of spatially varying coefficients, we employ the discovery-embedding process a single time. At this juncture, sparse regression effectively serves to furnish reasonable optimized initial coefficients and function terms for the next RCNN model. However, in the extreme case of sparse regression, the lack of a reasonable initial guess may lead to subsequent failures in RCNN models. We posit that the robust frameworks continuously proposed by the research community will mitigate this issue, and we intend to explore this further in future studies.

This study offers a conceptual comparison between viscoelastic wave physical systems and standard RNN. However, the design philosophy of our framework is universal. By implementing straightforward modifications to components such as recurrence relations, initial/boundary conditions, the framework can be adapted to address various types of PDEs. A central focus of our future research will be to develop a unified framework that is applicable across different classes of differential equations and/or applications, thereby further showcasing the inherent discovery capabilities embedded within this approach.

Furthermore, in recent years, numerous researchers have leveraged scientific machine learning (SciML) to embed PDEs into neural networks. However, the absence of governing equations can present significant challenges in practical scenarios. Our proposed hybrid approach has the potential to facilitate the discovery of more meaningful PDEs relevant to real-world scenarios. For example, it can assist in directly identifying unknown equations and inferring material properties from seismic data collected through experiments and observations. These equations can subsequently be integrated with the emerging SciML framework, utilizing established knowledge embedding techniques to create a closed-loop system of discovery and knowledge generation.

\section{Conclusion}\label{sec:conclusion}
We proposed a hybrid framework for the discovery and inversion of viscoelastic wave equation in inhomogeneous media, which couples the sparse regression technique and well-designed RCNN model. Our innovative paradigm leverages the manifold of priors derived from the techniques of discovering hidden PDEs from sparse data, the computational efficiency of numerical solvers, and well-established optimization tools. The proposed coupling scheme allows for the refinement of both the structure and coefficients of PDEs initially found by sparse regression using RCNN. As a result, the constraint of scalar coefficients identified by previous sparse discovery methods is overcome, enabling successful identification and inversion of wave propagation in heterogeneous media. In summary, we have presented an effective, interpretable, and flexible method for accurately and reliably discovering wave equations from imperfect, sparse, and noisy observational data.

\section*{Data availability}
All the datasets and source codes to reproduce the results in this study are available on GitHub at \url{https://github.com/norery/discovery_waveEQ} upon final publication.

\section*{Acknowledgements}
The authors are grateful to the editor and the reviewers for their constructive comments. The authors would like to thank the support by the research funding provided by the National Natural Science Foundation of China (NSFC, Grant Nos. 52192675, U1839202).

\printcredits

\bibliographystyle{elsarticle-num}

\bibliography{wave_discovery-bib.bib}

\begin{thebibliography}{10}
\expandafter\ifx\csname url\endcsname\relax
  \def\url#1{\texttt{#1}}\fi
\expandafter\ifx\csname urlprefix\endcsname\relax\def\urlprefix{URL }\fi
\expandafter\ifx\csname href\endcsname\relax
  \def\href#1#2{#2} \def\path#1{#1}\fi

\bibitem{Xu_2023_DiscoveryPartialDifferential}
H.~Xu, J.~Zeng, D.~Zhang, Discovery of {{Partial Differential Equations}} from
  {{Highly Noisy}} and {{Sparse Data}} with {{Physics-Informed Information
  Criterion}}, Research 6 (2023) 0147.
\newblock \href {https://doi.org/10.34133/research.0147}
  {\path{doi:10.34133/research.0147}}.

\bibitem{Yu_2024_LearningDynamicalSystems}
R.~Yu, R.~Wang, Learning dynamical systems from data: {{An}} introduction to
  physics-guided deep learning, Proceedings of the National Academy of Sciences
  121~(27) (2024) e2311808121.
\newblock \href {https://doi.org/10.1073/pnas.2311808121}
  {\path{doi:10.1073/pnas.2311808121}}.

\bibitem{Camps-Valls_2023_DiscoveringCausalRelations}
G.~{Camps-Valls}, A.~Gerhardus, U.~Ninad, G.~Varando, G.~Martius,
  E.~{Balaguer-Ballester}, R.~Vinuesa, E.~Diaz, L.~Zanna, J.~Runge, Discovering
  causal relations and equations from data, Physics Reports 1044 (2023) 1--68.
\newblock \href {https://doi.org/10.1016/j.physrep.2023.10.005}
  {\path{doi:10.1016/j.physrep.2023.10.005}}.

\bibitem{Bongard_2007_AutomatedReverseEngineering}
J.~Bongard, H.~Lipson, Automated reverse engineering of nonlinear dynamical
  systems, Proceedings of the National Academy of Sciences 104~(24) (2007)
  9943--9948.
\newblock \href {https://doi.org/10.1073/pnas.0609476104}
  {\path{doi:10.1073/pnas.0609476104}}.

\bibitem{Schmidt_2009_DistillingFreeFormNatural}
M.~Schmidt, H.~Lipson, Distilling {{Free-Form Natural Laws}} from
  {{Experimental Data}}, Science 324~(5923) (2009) 81--85.
\newblock \href {https://doi.org/10.1126/science.1165893}
  {\path{doi:10.1126/science.1165893}}.

\bibitem{Xu_2020_DLGAPDEDiscoveryPDEs}
H.~Xu, H.~Chang, D.~Zhang, {{DLGA-PDE}}: {{Discovery}} of {{PDEs}} with
  incomplete candidate library via combination of deep learning and genetic
  algorithm, Journal of Computational Physics 418 (2020) 109584.
\newblock \href {https://doi.org/10.1016/j.jcp.2020.109584}
  {\path{doi:10.1016/j.jcp.2020.109584}}.

\bibitem{Chen_2022_SymbolicGeneticAlgorithm}
Y.~Chen, Y.~Luo, Q.~Liu, H.~Xu, D.~Zhang, Symbolic genetic algorithm for
  discovering open-form partial differential equations ({{SGA-PDE}}), Physical
  Review Research 4~(2) (2022) 023174.
\newblock \href {https://doi.org/10.1103/PhysRevResearch.4.023174}
  {\path{doi:10.1103/PhysRevResearch.4.023174}}.

\bibitem{Makke_2024_InterpretableScientificDiscovery}
N.~Makke, S.~Chawla, Interpretable scientific discovery with symbolic
  regression: A review, Artificial Intelligence Review 57~(1) (2024) 2.
\newblock \href {https://doi.org/10.1007/s10462-023-10622-0}
  {\path{doi:10.1007/s10462-023-10622-0}}.

\bibitem{Cheng_2023_RobustDataDriven}
S.~Cheng, T.~Alkhalifah, Robust data driven discovery of a seismic wave
  equation, Geophysical Journal International 236~(1) (2023) 537--546.
\newblock \href {https://doi.org/10.1093/gji/ggad446}
  {\path{doi:10.1093/gji/ggad446}}.

\bibitem{Cheng_2024_DiscoveryPhysicallyInterpretable}
S.~Cheng, T.~Alkhalifah, Discovery of physically interpretable wave equations,
  arXiv preprint arXiv:2404.17971 (2024).

\bibitem{Brunton_2016_DiscoveringGoverningEquations}
S.~L. Brunton, J.~L. Proctor, J.~N. Kutz, Discovering governing equations from
  data by sparse identification of nonlinear dynamical systems, Proceedings of
  the National Academy of Sciences 113~(15) (2016) 3932--3937.
\newblock \href {https://doi.org/10.1073/pnas.1517384113}
  {\path{doi:10.1073/pnas.1517384113}}.

\bibitem{Loiseau_2018_SparseReducedorderModelling}
J.-C. Loiseau, B.~R. Noack, S.~L. Brunton, Sparse reduced-order modelling:
  Sensor-based dynamics to full-state estimation, Journal of Fluid Mechanics
  844 (2018) 459--490.
\newblock \href {https://doi.org/10.1017/jfm.2018.147}
  {\path{doi:10.1017/jfm.2018.147}}.

\bibitem{Loiseau_2018_ConstrainedSparseGalerkin}
J.-C. Loiseau, S.~L. Brunton, Constrained sparse {{Galerkin}} regression,
  Journal of Fluid Mechanics 838 (2018) 42--67.
\newblock \href {https://doi.org/10.1017/jfm.2017.823}
  {\path{doi:10.1017/jfm.2017.823}}.

\bibitem{Stender_2019_RecoveryDifferentialEquations}
M.~Stender, S.~Oberst, N.~Hoffmann, Recovery of {{Differential Equations}} from
  {{Impulse Response Time Series Data}} for {{Model Identification}} and
  {{Feature Extraction}}, Vibration 2~(1) (2019) 25--46.
\newblock \href {https://doi.org/10.3390/vibration2010002}
  {\path{doi:10.3390/vibration2010002}}.

\bibitem{Rudy_2017_DatadrivenDiscoveryPartial}
S.~H. Rudy, S.~L. Brunton, J.~L. Proctor, J.~N. Kutz, Data-driven discovery of
  partial differential equations, Science Advances 3~(4) (2017) e1602614.
\newblock \href {https://doi.org/10.1126/sciadv.1602614}
  {\path{doi:10.1126/sciadv.1602614}}.

\bibitem{Messenger_2021_WeakSINDyPartial}
D.~A. Messenger, D.~M. Bortz, Weak {{SINDy}} for partial differential
  equations, Journal of Computational Physics 443 (2021) 110525.
\newblock \href {https://doi.org/10.1016/j.jcp.2021.110525}
  {\path{doi:10.1016/j.jcp.2021.110525}}.

\bibitem{Sashidhar_2022_BaggingOptimizedDynamic}
D.~Sashidhar, J.~N. Kutz, Bagging, optimized dynamic mode decomposition for
  robust, stable forecasting with spatial and temporal uncertainty
  quantification, Philosophical Transactions of the Royal Society A:
  Mathematical, Physical and Engineering Sciences 380~(2229) (2022) 20210199.
\newblock \href {https://doi.org/10.1098/rsta.2021.0199}
  {\path{doi:10.1098/rsta.2021.0199}}.

\bibitem{Fasel_2022_EnsembleSINDyRobustSparse}
U.~Fasel, J.~N. Kutz, B.~W. Brunton, S.~L. Brunton, Ensemble-{{SINDy}}:
  {{Robust}} sparse model discovery in the low-data, high-noise limit, with
  active learning and control, Proceedings of the Royal Society A:
  Mathematical, Physical and Engineering Sciences 478~(2260) (2022) 20210904.
\newblock \href {https://doi.org/10.1098/rspa.2021.0904}
  {\path{doi:10.1098/rspa.2021.0904}}.

\bibitem{Chen_2021_PhysicsinformedLearningGoverning}
Z.~Chen, Y.~Liu, H.~Sun, Physics-informed learning of governing equations from
  scarce data, Nature Communications 12~(1) (2021) 6136.
\newblock \href {https://doi.org/10.1038/s41467-021-26434-1}
  {\path{doi:10.1038/s41467-021-26434-1}}.

\bibitem{Rao_2023_EncodingPhysicsLearn}
C.~Rao, P.~Ren, Q.~Wang, O.~Buyukozturk, H.~Sun, Y.~Liu, Encoding physics to
  learn reaction--diffusion processes, Nature Machine Intelligence 5~(7) (2023)
  765--779.
\newblock \href {https://doi.org/10.1038/s42256-023-00685-7}
  {\path{doi:10.1038/s42256-023-00685-7}}.

\bibitem{Raissi_2019_PhysicsinformedNeuralNetworks}
M.~Raissi, P.~Perdikaris, G.~Karniadakis, Physics-informed neural networks:
  {{A}} deep learning framework for solving forward and inverse problems
  involving nonlinear partial differential equations, Journal of Computational
  Physics 378 (2019) 686--707.
\newblock \href {https://doi.org/10.1016/j.jcp.2018.10.045}
  {\path{doi:10.1016/j.jcp.2018.10.045}}.

\bibitem{Song_2021_SolvingFrequencydomainAcoustic}
C.~Song, T.~Alkhalifah, U.~B. Waheed, Solving the frequency-domain acoustic
  {{VTI}} wave equation using physics-informed neural networks, Geophysical
  Journal International 225~(2) (2021) 846--859.
\newblock \href {https://doi.org/10.1093/gji/ggab010}
  {\path{doi:10.1093/gji/ggab010}}.

\bibitem{Ding_2023_SelfadaptivePhysicsdrivenDeep}
Y.~Ding, S.~Chen, X.~Li, S.~Wang, S.~Luan, H.~Sun, Self-adaptive physics-driven
  deep learning for seismic wave modeling in complex topography, Engineering
  Applications of Artificial Intelligence 123 (2023) 106425.
\newblock \href {https://doi.org/10.1016/j.engappai.2023.106425}
  {\path{doi:10.1016/j.engappai.2023.106425}}.

\bibitem{Ding_2023_PhysicsconstrainedNeuralNetworks}
Y.~Ding, S.~Chen, X.~Li, L.~Jin, S.~Luan, H.~Sun, Physics-constrained neural
  networks for half-space seismic wave modeling, Computers \& Geosciences 181
  (2023) 105477.
\newblock \href {https://doi.org/10.1016/j.cageo.2023.105477}
  {\path{doi:10.1016/j.cageo.2023.105477}}.

\bibitem{Ren_2024_SeismicNetPhysicsinformedNeural}
P.~Ren, C.~Rao, S.~Chen, J.-X. Wang, H.~Sun, Y.~Liu, {{SeismicNet}}:
  {{Physics-informed}} neural networks for seismic wave modeling in
  semi-infinite domain, Computer Physics Communications 295 (2024) 109010.
\newblock \href {https://doi.org/10.1016/j.cpc.2023.109010}
  {\path{doi:10.1016/j.cpc.2023.109010}}.

\bibitem{Rasht-Behesht_2022_PhysicsInformedNeuralNetworks}
M.~Rasht-Behesht, C.~Huber, K.~Shukla, G.~E. Karniadakis, Physics-{{Informed
  Neural Networks}} ({{PINNs}}) for {{Wave Propagation}} and {{Full Waveform
  Inversions}}, Journal of Geophysical Research: Solid Earth 127~(5) (2022)
  e2021JB023120.
\newblock \href {https://doi.org/10.1029/2021JB023120}
  {\path{doi:10.1029/2021JB023120}}.

\bibitem{Zhang_2023_SeismicInversionBased}
Y.~Zhang, X.~Zhu, J.~Gao, Seismic {{Inversion Based}} on {{Acoustic Wave
  Equations Using Physics-Informed Neural Network}}, IEEE Transactions on
  Geoscience and Remote Sensing 61 (2023) 1--11.
\newblock \href {https://doi.org/10.1109/TGRS.2023.3236973}
  {\path{doi:10.1109/TGRS.2023.3236973}}.

\bibitem{Gao_2021_PhyGeoNetPhysicsinformedGeometryadaptive}
H.~Gao, L.~Sun, J.-X. Wang, {{PhyGeoNet}}: {{Physics-informed}}
  geometry-adaptive convolutional neural networks for solving parameterized
  steady-state {{PDEs}} on irregular domain, Journal of Computational Physics
  428 (2021) 110079.
\newblock \href {https://doi.org/10.1016/j.jcp.2020.110079}
  {\path{doi:10.1016/j.jcp.2020.110079}}.

\bibitem{Qu_2022_LearningTimedependentPDEs}
J.~Qu, W.~Cai, Y.~Zhao, Learning time-dependent {{PDEs}} with a linear and
  nonlinear separate convolutional neural network, Journal of Computational
  Physics 453 (2022) 110928.
\newblock \href {https://doi.org/10.1016/j.jcp.2021.110928}
  {\path{doi:10.1016/j.jcp.2021.110928}}.

\bibitem{Ren_2022_PhyCRNetPhysicsinformedConvolutionalrecurrent}
P.~Ren, C.~Rao, Y.~Liu, J.-X. Wang, H.~Sun, {{PhyCRNet}}: {{Physics-informed}}
  convolutional-recurrent network for solving spatiotemporal {{PDEs}}, Computer
  Methods in Applied Mechanics and Engineering 389 (2022) 114399.
\newblock \href {https://doi.org/10.1016/j.cma.2021.114399}
  {\path{doi:10.1016/j.cma.2021.114399}}.

\bibitem{McGreivy_2024_WeakBaselinesReporting}
N.~McGreivy, A.~Hakim, Weak baselines and reporting biases lead to overoptimism
  in machine learning for fluid-related partial differential equations, arXiv
  preprint arXiv:2407.07218 (2024).

\bibitem{Tancik_2020_FourierFeaturesLet}
M.~Tancik, P.~Srinivasan, B.~Mildenhall, S.~{Fridovich-Keil}, N.~Raghavan,
  U.~Singhal, R.~Ramamoorthi, J.~Barron, R.~Ng, Fourier features let networks
  learn high frequency functions in low dimensional domains, Advances in Neural
  Information Processing Systems 33 (2020) 7537--7547.

\bibitem{Wang_2021_EigenvectorBiasFourier}
S.~Wang, H.~Wang, P.~Perdikaris, On the eigenvector bias of {{Fourier}} feature
  networks: {{From}} regression to solving multi-scale {{PDEs}} with
  physics-informed neural networks, Computer Methods in Applied Mechanics and
  Engineering 384 (2021) 113938.
\newblock \href {http://arxiv.org/abs/2012.10047} {\path{arXiv:2012.10047}},
  \href {https://doi.org/10.1016/j.cma.2021.113938}
  {\path{doi:10.1016/j.cma.2021.113938}}.

\bibitem{Guo_2023_ElectromagneticModelingUsing}
L.~Guo, M.~Li, S.~Xu, F.~Yang, L.~Liu, Electromagnetic {{Modeling Using}} an
  {{FDTD-Equivalent Recurrent Convolution Neural Network}}: {{Accurate}}
  computing on a deep learning framework, IEEE Antennas and Propagation
  Magazine 65~(1) (2023) 93--102.
\newblock \href {https://doi.org/10.1109/MAP.2021.3127514}
  {\path{doi:10.1109/MAP.2021.3127514}}.

\bibitem{Ji_2024_EfficientPlatformNumerical}
D.~Ji, C.~Li, C.~Zhai, Z.~Cao, An {{Efficient Platform}} for {{Numerical
  Modeling}} of {{Partial Differential Equations}}, IEEE Transactions on
  Geoscience and Remote Sensing 62 (2024) 1--13.
\newblock \href {https://doi.org/10.1109/TGRS.2024.3409620}
  {\path{doi:10.1109/TGRS.2024.3409620}}.

\bibitem{Idriss_1968_SeismicResponseHorizontal}
I.~M. Idriss, H.~B. Seed, Seismic {{Response}} of {{Horizontal Soil Layers}},
  Journal of the Soil Mechanics and Foundations Division 94~(4) (1968)
  1003--1031.
\newblock \href {https://doi.org/10.1061/JSFEAQ.0001163}
  {\path{doi:10.1061/JSFEAQ.0001163}}.

\bibitem{Tibshirani_1996_RegressionShrinkageSelection}
R.~Tibshirani, Regression {{Shrinkage}} and {{Selection}} via the {{Lasso}},
  Journal of the Royal Statistical Society Series B: Statistical Methodology
  58~(1) (1996) 267--288.
\newblock \href {http://arxiv.org/abs/2346178} {\path{arXiv:2346178}}, \href
  {https://doi.org/10.1111/j.2517-6161.1996.tb02080.x}
  {\path{doi:10.1111/j.2517-6161.1996.tb02080.x}}.

\bibitem{Hughes_2019_WavePhysicsAnalog}
T.~W. Hughes, I.~A.~D. Williamson, M.~Minkov, S.~Fan, Wave physics as an analog
  recurrent neural network, Science Advances 5~(12) (2019) eaay6946.
\newblock \href {https://doi.org/10.1126/sciadv.aay6946}
  {\path{doi:10.1126/sciadv.aay6946}}.

\bibitem{Wu_2023_NeuralPartialDifferential}
Z.~Wu, X.~He, Y.~Li, C.~Yang, R.~Liu, S.~Xiong, B.~Zhu, Neural partial
  differential equations with functional convolution, arXiv preprint
  arXiv:2303.07194 (2023).

\bibitem{LeCun_1995_ConvolutionalNetworksImages}
Y.~LeCun, Y.~Bengio, Convolutional networks for images, speech, and time
  series, The handbook of brain theory and neural networks 3361~(10) (1995)
  1995.

\bibitem{Cai_2012_ImageRestorationTotal}
J.-F. Cai, B.~Dong, S.~Osher, Z.~Shen, Image restoration: {{Total}} variation,
  wavelet frames, and beyond, Journal of the American Mathematical Society
  25~(4) (2012) 1033--1089.
\newblock \href {https://doi.org/10.1090/S0894-0347-2012-00740-1}
  {\path{doi:10.1090/S0894-0347-2012-00740-1}}.

\bibitem{Dong_2017_ImageRestorationWavelet}
B.~Dong, Q.~Jiang, Z.~Shen, Image {{Restoration}}: {{Wavelet Frame Shrinkage}},
  {{Nonlinear Evolution PDEs}}, and {{Beyond}}, Multiscale Modeling \&
  Simulation 15~(1) (2017) 606--660.
\newblock \href {https://doi.org/10.1137/15M1037457}
  {\path{doi:10.1137/15M1037457}}.

\bibitem{Kingma_2014_AdamMethodStochastic}
D.~P. Kingma, Adam: A method for stochastic optimization, arXiv preprint
  arXiv:1412.6980 (2014).

\bibitem{Liu_1989_LimitedMemoryBFGS}
D.~C. Liu, J.~Nocedal, On the limited memory {{BFGS}} method for large scale
  optimization, Mathematical Programming 45~(1-3) (1989) 503--528.
\newblock \href {https://doi.org/10.1007/BF01589116}
  {\path{doi:10.1007/BF01589116}}.

\bibitem{Wang_2019_FourthOrderFinite}
S.~Wang, N.~A. Petersson, Fourth {{Order Finite Difference Methods}} for the
  {{Wave Equation}} with {{Mesh Refinement Interfaces}}, SIAM Journal on
  Scientific Computing 41~(5) (2019) A3246--A3275.
\newblock \href {https://doi.org/10.1137/18M1211465}
  {\path{doi:10.1137/18M1211465}}.

\bibitem{Liao_1984_TransmittingBoundaryNumerical}
Z.~Liao, H.~Wong, A transmitting boundary for the numerical simulation of
  elastic wave propagation, International Journal of Soil Dynamics and
  Earthquake Engineering 3~(4) (1984) 174--183.
\newblock \href {https://doi.org/10.1016/0261-7277(84)90033-0}
  {\path{doi:10.1016/0261-7277(84)90033-0}}.

\bibitem{Xing_2021_TheoryNewUnified}
H.~Xing, X.~Li, H.~Li, Z.~Xie, S.~Chen, Z.~Zhou, The {{Theory}} and {{New
  Unified Formulas}} of {{Displacement-Type Local Absorbing Boundary
  Conditions}}, Bulletin of the Seismological Society of America 111~(2) (2021)
  801--824.
\newblock \href {https://doi.org/10.1785/0120200155}
  {\path{doi:10.1785/0120200155}}.

\bibitem{Xing_2021_SpectralelementFormulationMultitransmitting}
H.~Xing, X.~Li, H.~Li, A.~Liu, Spectral-element formulation of
  multi-transmitting formula and its accuracy and stability in {{1D}} and
  {{2D}} seismic wave modeling, Soil Dynamics and Earthquake Engineering 140
  (2021) 106218.
\newblock \href {https://doi.org/10.1016/j.soildyn.2020.106218}
  {\path{doi:10.1016/j.soildyn.2020.106218}}.

\bibitem{Liao_1992_NumericalInstabilitiesLocal}
Z.~P. Liao, J.~B. Liu, Numerical instabilities of a local transmitting
  boundary, Earthquake Engineering \& Structural Dynamics 21~(1) (1992) 65--77.
\newblock \href {https://doi.org/10.1002/eqe.4290210105}
  {\path{doi:10.1002/eqe.4290210105}}.

\bibitem{Charlier_1986_WaveEquationsLinear}
J.-P. Charlier, F.~Crowet, Wave equations in linear viscoelastic materials, The
  Journal of the Acoustical Society of America 79~(4) (1986) 895--900.
\newblock \href {https://doi.org/10.1121/1.393685}
  {\path{doi:10.1121/1.393685}}.

\bibitem{Paszke_2019_PyTorchImperativeStylea}
A.~Paszke, S.~Gross, F.~Massa, A.~Lerer, J.~Bradbury, G.~Chanan, T.~Killeen,
  Z.~Lin, N.~Gimelshein, L.~Antiga, Pytorch: {{An}} imperative style,
  high-performance deep learning library, Advances in neural information
  processing systems 32 (2019).

\bibitem{Udrescu_2020_AIFeynmanPhysicsinspired}
S.-M. Udrescu, M.~Tegmark, {{AI Feynman}}: {{A}} physics-inspired method for
  symbolic regression, Science Advances 6~(16) (2020) eaay2631.
\newblock \href {https://doi.org/10.1126/sciadv.aay2631}
  {\path{doi:10.1126/sciadv.aay2631}}.

\bibitem{Buckingham_1914_PhysicallySimilarSystems}
E.~Buckingham, On {{Physically Similar Systems}}; {{Illustrations}} of the
  {{Use}} of {{Dimensional Equations}}, Physical Review 4~(4) (1914) 345--376.
\newblock \href {https://doi.org/10.1103/PhysRev.4.345}
  {\path{doi:10.1103/PhysRev.4.345}}.

\bibitem{Matchev_2022_AnalyticalModelingExoplanet}
K.~T. Matchev, K.~Matcheva, A.~Roman, Analytical {{Modeling}} of {{Exoplanet
  Transit Spectroscopy}} with {{Dimensional Analysis}} and {{Symbolic
  Regression}}, The Astrophysical Journal 930~(1) (2022) 33.
\newblock \href {https://doi.org/10.3847/1538-4357/ac610c}
  {\path{doi:10.3847/1538-4357/ac610c}}.

\bibitem{Tenachi_2023_DeepSymbolicRegression}
W.~Tenachi, R.~Ibata, F.~I. Diakogiannis, Deep {{Symbolic Regression}} for
  {{Physics Guided}} by {{Units Constraints}}: {{Toward}} the {{Automated
  Discovery}} of {{Physical Laws}}, The Astrophysical Journal 959~(2) (2023)
  99.
\newblock \href {https://doi.org/10.3847/1538-4357/ad014c}
  {\path{doi:10.3847/1538-4357/ad014c}}.

\end{thebibliography}

\end{document}